\documentclass[runningheads]{llncs}
\usepackage{amssymb}
\usepackage{amsmath}
\usepackage{graphicx}
\usepackage{todonotes}
\usepackage{booktabs}
\usepackage[english]{babel}
\usepackage{float}
\usepackage[utf8x]{inputenc}
\usepackage{calc}
\usepackage{verbatim}

\usepackage{graphicx,placeins,lscape,
  colortbl,longtable,footmisc,
  multirow}
\usepackage{hyperref}

\definecolor{lime}{HTML}{A6CE39}
\DeclareRobustCommand{\orcidicon}{%
	\begin{tikzpicture}
	\draw[lime, fill=lime] (0,0) 
	circle [radius=0.16] 
	node[white] {{\fontfamily{qag}\selectfont \tiny ID}};
	\draw[white, fill=white] (-0.0625,0.095) 
	circle [radius=0.007];
	\end{tikzpicture}
	\hspace{-2mm}
}

\foreach \x in {A, ..., Z}{%
	\expandafter\xdef\csname orcid\x\endcsname{\noexpand\href{https://orcid.org/\csname orcidauthor\x\endcsname}{\noexpand\orcidicon}}
}

\newcommand{\magic}[0]{XNAP} %

\begin{document}
\title{XNAP: Making LSTM-based Next Activity Predictions Explainable by Using LRP}
\titlerunning{XNAP: Explainable next activity prediction}
\author{Sven Weinzierl\inst{1}\orcidA{} \and
Sandra Zilker\inst{1}\orcidB{} \and
Jens Brunk\inst{2}\orcidC{} \and
Kate Revoredo\inst{3}\orcidD{} \and
Martin Matzner\inst{1}\orcidE{} \and
Jörg Becker\inst{2}\orcidF{}
}

\authorrunning{S. Weinzierl et al.}
\institute{
Institute of Information Systems, Friedrich-Alexander-Universit{\"a}t Erlangen-N{\"u}rnberg, F{\"u}rther Stra{\ss}e 248, Germany \and 
European Research Center for Information Systems (ERCIS), University of Münster, Leonardo-Campus 3, 48149 M{\"u}nster \and
Department of Information Systems and Operations, Vienna University of Economics and Business (WU), Vienna, Austria \\
\email{\{sven.weinzierl, sandra.zilker, martin.matzner\}@fau.de}\\
\email{\{jens.brunk, becker\}@ercis.uni-muenster.de}\\
\email{kate.revoredo@wu.ac.at}
}

\maketitle              %
\begin{abstract}
Predictive business process monitoring (PBPM) is a class of techniques designed to predict behaviour, such as next activities, in running traces. 
PBPM techniques aim to improve process performance by providing predictions to process analysts, supporting them in their decision making. 
However, the PBPM techniques' limited predictive quality was considered as the essential obstacle for establishing such techniques in practice. 
With the use of deep neural networks (DNNs), the techniques' predictive quality could be improved for tasks like the next activity prediction.
While DNNs achieve a promising predictive quality, they still lack comprehensibility due to their hierarchical approach of learning representations. 
Nevertheless, process analysts need to comprehend the cause of a prediction to identify intervention mechanisms that might affect the decision making to secure process performance. 
In this paper, we propose \magic{}, the first explainable, DNN-based PBPM technique for the next activity prediction. %
\magic{} integrates a layer-wise relevance propagation method from the field of explainable artificial intelligence to make predictions of a long short-term memory DNN explainable by providing relevance values for activities. We show the benefit of our approach through two real-life event logs. 
\end{abstract}
\section{Introduction}
Predictive business process monitoring (PBPM)~\cite{maggi.2014} emerged in the field of business process management (BPM) to improve the performance of operational business processes~\cite{breuker.2016,schwegmann-2013}.
PBPM is a class of techniques designed to predict behaviour, such as next activities, in running traces. 
PBPM techniques aim to improve process performance by providing predictions to process analysts, supporting them in their decision making.
Predictions may reveal inefficiencies, risks and mistakes in traces supporting process analysts on their decisions to mitigate the issues~\cite{di.2017}.
Typically, PBPM techniques use predictive models, that are extracted from historical event log data. Most of the current techniques apply ``traditional" machine-learning (ML) algorithms to learn models, which produce predictions with a higher predictive quality~\cite{di.2018}. 
The PBPM techniques' limited predictive quality was considered as the essential obstacle for establishing such techniques in practice~\cite{weinzierl.2019}. Therefore, a plethora of works has proposed approaches to further increase predictive quality~\cite{sindhgatta.2019}.
By using deep neural networks (DNNs), the techniques' predictive quality was improved for tasks like the next activity prediction~\cite{evermann.2016}. 

\begin{figure}[ht]
\centering
\includegraphics[width=9cm]{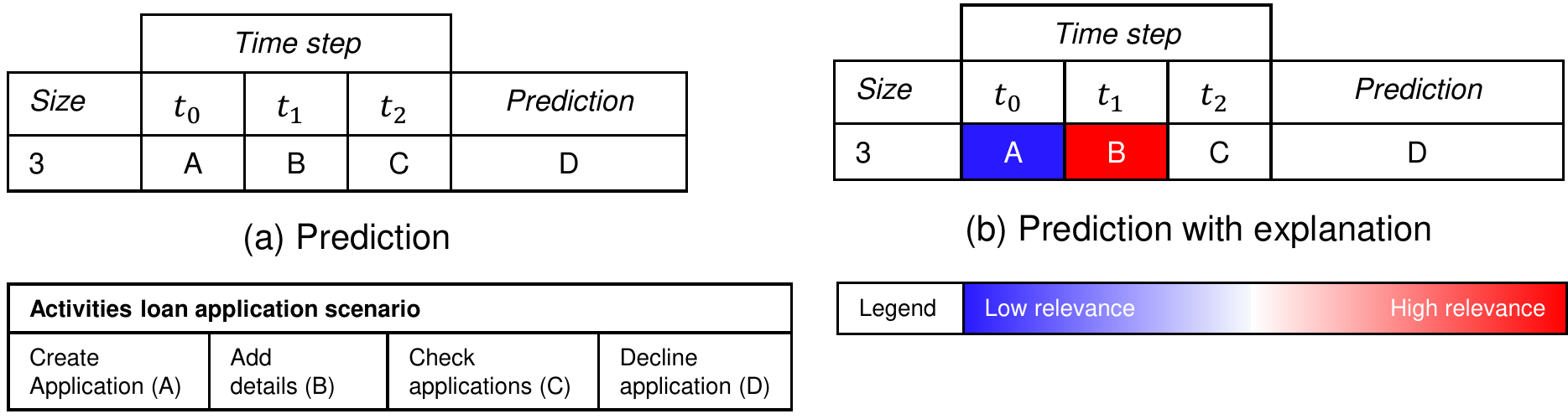}
\caption{Next activity prediction example without (a) and with explanation (b).} 
\label{fig:example}
\end{figure}

In practice, a process analyst's choice to use a PBPM technique does not only depend on a PBPM technique's predictive quality. 
Márquez-Chamorro et al.~\cite{marquez.2017a} state that the explainability of a PBPM technique's predictions is also an important factor for using such a technique in practice.
By providing an explanation of a prediction, the process analyst's confidence in a PBPM technique improves and the process analyst may adopt the PBPM technique~\cite{nunes.2017}.  
However, DNNs learn multiple representations to find the intricate structure in data, and therefore the cause of a prediction is difficult to retrieve~\cite{lecun.2015}. 
Due to the lack of explainability, a process analysts cannot identify intervention mechanisms that might affect the decision making to secure the process performance. 
To address this issue, explainable artificial intelligence (XAI) has developed as a sub-field of artificial intelligence.
XAI is a class of ML techniques that aims to enable humans to understand, trust and manage the advanced artificial ``decision-supporters" by producing more explainable models, while maintaining a high level of predictive quality~\cite{gunning.2017}.   
For instance, in a loan application process, the prediction of the next activity ``Decline application" (cf. (a) in Fig.~\ref{fig:example}) produced by a model trained with a DNN can be insufficient for a process analyst to decide if this is a normal behaviour or some intervention is required to avoid an unnecessary refusal of the application.
In contrast, the prediction with explanation (cf. (b) in Fig.~\ref{fig:example}) informs the process analyst that some important details are missing for approving the application because the activity ``Add details" has a high relevance on the prediction of the next activity ``Decline application".

\vspace{-0.5cm}
\vspace{-0.5cm}

In this paper, we propose the explainable PBPM technique \magic. \magic{} integrates a layer-wise relevance propagation (LRP) method from XAI to make next activity predictions of a long short-term memory (LSTM) DNN explainable by providing relevance values for each activity in the course of a running trace.
To the best of the authors' knowledge, this work proposes the first approach to make LSTM-based next activity predictions explainable.

The paper is structured as follows. 
Sec.~\ref{sec:background} introduces the required background. 
In Sec.~\ref{sec:relatedWork}, we present related work on explainable PBPM and reveal the research gap. 
Sec.~\ref{sec:xnap} introduces the design of \magic. 
In Sec.~\ref{sec:eval}, the benefits of \magic{} are demonstrated based on two real-life event logs. 
In Sec.~\ref{sec:conclusion} we provide a summary and point to future research directions.

\section{Background}
\label{sec:background}
\subsection{Preliminaries\protect\footnote{Note definitions are inspired by the work of Taymouri et al.~\cite{taymouri.2020}.}}
\begin{definition}[Vector, Matrix, Tensor]%
A vector $\mathbf{x}=\left(x_{1}, x_{2}, \ldots, x_{n}\right)^{T}$ is an array of numbers, in which the i\textsuperscript{th} number is identified by $\mathbf{x}_{i}$. If each number of vector $\mathbf{x}$ lies in $\mathbb{R}$ and the vector $\mathbf{x}$ contains $n$ numbers, then the vector $\mathbf{x}$ lies in $\mathbb{R}^{n \times 1}$, and the vector $\mathbf{x}$'s dimension is $n \times 1$.
A matrix $\mathbf{M}=\left(\mathbf{x}^{(1)}, \mathbf{x}^{(2)}, \ldots, \mathbf{x}^{(n)}\right)$ is a two-dimensional array of numbers, where $\mathbf{M} \in \mathbb{R}^{n \times d}.$ %
A tensor $\mathsf{T}$ is an $n$-dimensional array of numbers. If $n=3$, then $\mathsf{T}$ is a tensor of the third order with $\mathsf{T}=\left(\mathbf{M}^{(1)}, \mathbf{M}^{(2)}, \ldots, \mathbf{M}^{(n)}\right)$,
where $\mathsf{T} \in \mathbb{R}^{n \times b \times u}.$ %
\end{definition}

\begin{definition}[Event, Trace, Event Log] An event is a tuple $(c,a,t)$ where $c$ is the case id, $a$ is the activity (event type) and $t$ is the timestamp. A trace is a non-empty sequence
$\sigma=\left\langle e_{1}, \ldots, e_{\vert \sigma \vert} \right\rangle$ of events such that $\forall i, j \in\{1, \ldots, \vert \sigma \vert\}$ $e_{i} . c=e_{j}.c.$ An event log $L$ is
a set $\left\{\sigma_{1}, \ldots, \sigma_{\vert L \vert}\right\}$ of traces.
A trace %
can also be considered as a sequence of vectors, in which a vector contains all or a part of the information relating to an event, e.g. an event's activity. Formally, $\sigma=\left\langle\mathbf{x}^{(1)}, \mathbf{x}^{(2)}, \ldots, \mathbf{x}^{(t)}\right\rangle,$ where $\mathbf{x}^{(i)} \in \mathbb{R}^{{n} \times 1}$ is a vector, and the
superscript indicates the time-order upon which the events happened.
\end{definition}

\begin{definition}[Prefix and label] Given a trace $\sigma=\left\langle e_{1},\dots, e_{k}, \dots, e_{\vert \sigma \vert }\right\rangle$, a prefix of length $k$, that is a non-empty sequence, is defined as $f_{p}^{(k)}(\sigma)=\langle e_{1},\dots, e_{k}\rangle,$ with $0 < k < \vert \sigma_{c} \vert$ and a label (i.e. next activity) for a prefix of length $k$ is defined as $f_{l}^{(k)}(\sigma)=\langle e_{k+1}\rangle$.   
The above definition also holds for an input trace representing a sequence of vectors. For example, the tuple of all possible prefixes and the tuple of all possible labels for 
$\sigma=\langle\mathbf{x}^{(1)}, \mathbf{x}^{(2)}, \mathbf{x}^{(3)} 
\rangle$
are 
$\langle
\langle\mathbf{x}^{(1)}\rangle,
\langle\mathbf{x}^{(1)},\mathbf{x}^{(2)}\rangle
\rangle$
and
$\langle\mathbf{x}^{(2)},
\mathbf{x}^{(3)}
\rangle$.

\end{definition}

\subsection{Layer-wise Relevance Propagation for LSTMs}
LRP is a technique to explain predictions of DNNs in terms of input variables~\cite{bach.2015}.
For a given input sequence $\sigma = \langle \mathbf{x}^{(1)}, \mathbf{x}^{(2)}, \mathbf{x}^{(3)}\rangle$, a trained DNN model $\mathcal{M}_{c}$ and a calculated prediction $\mathbf{o}=\mathcal{M}_{c}(\sigma)$, 
LRP reverse-propagates the prediction $\mathbf{o}$ through the DNN model $\mathcal{M}_{c}$ to assign a relevance value to each input variable of $\sigma$~\cite{arras.2019}. 
A relevance value indicates to which extent an input variable contributes to the prediction. 
Note $\mathcal{M}_c$ is a DNN model, and $c$ is a target class for which we want to perform LRP. 
In this paper, $\mathcal{M}_c$ is an LSTM model, i.e. a DNN model with an LSTM~\cite{hochreiter.1997} layer as a hidden layer. The architecture of the ``vanilla" LSTM (layer) is common in the PBPM literature for the task of predicting next activities~\cite{weinzierl.2020}. For instance, an explanation of it can be found in the work of Evermann~et~al.~\cite{evermann.2016}. 

To calculate the relevance values of the input variables, LRP performs two computational steps.
First, it sets the relevance of an output layer neuron
corresponding to the target class of interest $c$
to the value $\mathbf{o}=\mathcal{M}_{c}(\sigma)$.
It ignores the other output layer neurons and equivalently sets their relevance to zero.
Second, it computes a relevance value
for each intermediate lower-layer neuron
depending on the neural connection type.
A DNN's layer can be described by one or more neural connections.
In turns, the LRP procedure can be described layer-by-layer for different types of layers included in a DNN. Depending on the type of a neural connection, LRP defines heuristic propagation rules for attributing the relevance to lower-layer neurons given the relevance values of the upper-layer neurons~\cite{bach.2015}. 

In case of recurrent neural network layers, such as LSTM~\cite{hochreiter.1997} layers,
there are two types of neural connections: \emph{many-to-one weighted linear connections}, and \emph{two-to-one multiplicative interactions}~\cite{arras.2017}. Therefore, we restrict the definition of the LRP procedure to these types of connections.
For weighted connections, let $\mathbf{z}_{j}$ be an upper-layer neuron.
Its value in the forward pass is computed as $\mathbf{z}_{j}=\sum_{i} \mathbf{z}_{i} \cdot \mathbf{w}_{ij} + b_{j}$,
while $\mathbf{z}_{i}$ are the lower-layer neurons,
and $\mathbf{w}_{ij}$ as well as $\mathbf{b}_{j}$ are the connection weights and biases.
Given each relevance $\mathbf{R}_{j}$ of the upper-layer neurons $\mathbf{z}_{j}$,
LRP computes the relevance $\mathbf{R}_{i}$ of the lower-layer neurons $\mathbf{z}_{i}$. Initially, $\mathbf{R}_{j}=\mathcal{M}_{c}(\sigma)$ is set. The relevance distribution onto lower-layer neurons comprises two steps. First, by computing relevance messages $\mathbf{R}_{i \leftarrow j}$ going from upper-layer neurons $\mathbf{z}_{j}$ to lower-layer neurons $\mathbf{z}_{i}$. The messages $\mathbf{R}_{i\leftarrow j}$ are computed as a fraction of the relevance $\mathbf{R}_{j}$ accordingly to the following rule:

\begin{equation}
\mathbf{R}_{i \leftarrow j}=\frac{\mathbf{z}_{i}\cdot \mathbf{w}_{ij}+\frac{\epsilon \cdot sign(\mathbf{z}_{j})+\delta \cdot \mathbf{b}_{j}}{N}}{\mathbf{z}_{j}+\epsilon \cdot sign(\mathbf{z}_{j})} \cdot \mathbf{R}_{j}.
\end{equation}

$N$ is the total number of lower-layer neurons connected to $\mathbf{z}_{j}$,
$\epsilon$ is a stabiliser~(small positive number, e.g. $0.001$) and $sign(\mathbf{z}_{j})=(1_{\mathbf{z}_{j}\geq 0} - 1_{\mathbf{z}_{j} < 0})$ is the sign of $\mathbf{z}_{j}$.
Second, by summing up incoming messages for each lower-layer neuron $\mathbf{z}_{i}$ to obtain relevance $\mathbf{R}_{i}$. $\mathbf{R}_{i}$ is computed as $\sum_{j}\mathbf{R}_{i\leftarrow j}$.
If the multiplicative factor $\delta$ is set to 1.0,
the total relevance of all neurons in the same layer is conserved.
If it is set to 0.0, the total relevance is absorbed by the biases.

For two-to-one multiplicative interactions between lower-layer neurons, let $\mathbf{z}_{j}$ be an upper-layer neuron.
Its value in the forward pass is computed as the multiplication of two lower-layer neuron values $\mathbf{z}_{g}$ and $\mathbf{z}_{s}$, i.e. $\mathbf{z}_{j}=\mathbf{z}_{g} \cdot \mathbf{z}_{s}$.
In such multiplicative interactions,
there is always one of two lower-layer neurons
that represents a gate with a value range~$[0,1]$
as the output of a sigmoid activation function.
This neuron is called gate $\mathbf{z}_{g}$, whereas the remaining one is the source $\mathbf{z}_{s}$.
Given such a configuration,
and denoting by $\mathbf{R}_{j}$ the relevance of the upper-layer neuron $\mathbf{z}_{j}$,
the relevance can be redistributed onto lower-layer neurons by: $\mathbf{R}_{g} = 0$ and $\mathbf{R}_{s} = \mathbf{R}_{j}$.
With this reallocation rule,
the gate neuron already decides
in the forward pass
how much of the information contained in the source neuron
should be retained to make the overall classification decision.

\section{Related Work on Explainable PBPM}
\label{sec:relatedWork}

In the past, PBPM research has mainly focus on improving the predictive quality of PBPM approaches to foster the transfer of these approaches into practice. In contrast, the PBPM approaches' explainability was scarcely discussed although it can be equally important since missing explainability might limit the PBPM approaches' applicability~\cite{marquez.2017a}. 
In the context of ML, XAI has already been considered in different approaches~\cite{BarredoArrieta2020}. 
However, PBPM research has just recently started to focus on XAI. %
Researchers differentiate between two types of explainability. First, ante-hoc explainability provides transparency on different levels of the model itself; thus they are referred to as transparent models. This can be the complete model, single components or learning algorithms. Second, post-hoc explainability can be provided in the form of visualisations after the model was trained since they are extracted from the trained model~\cite{du2019techniques}.%

Concerning ante-hoc explainability in PBPM, multiple approaches have been proposed for different prediction tasks.
For example, Maggi et al.~\cite{maggi.2014} propose a decision-tree-based, Breuker et al.~\cite{breuker.2016} a probabilistic-based, Rehse et al.~\cite{rehse.2019} a rule-based and Senderovic et al.~\cite{Senderovich2017} a regression-based approach.

\begin{figure}[htpb]
    \centering
    \includegraphics[width=11.0cm]{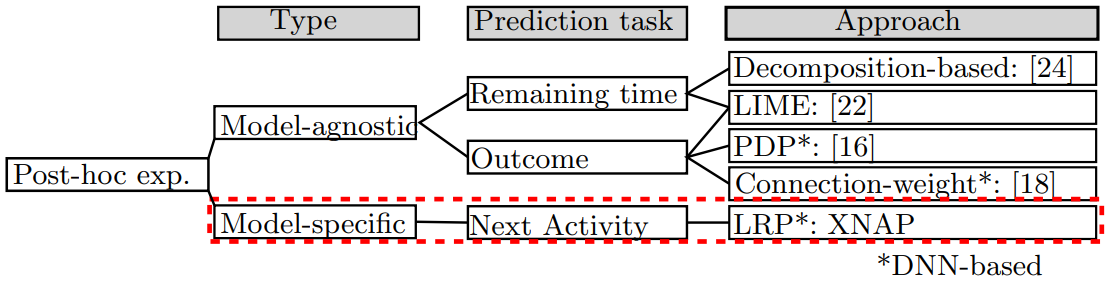}
    \caption{Related work on XAI in PBPM.}
    \label{fig:RelatedWork}
\end{figure}

In terms of post-hoc explainability, research has focused on model-agnostic approaches. These are techniques that can be added to any model in order to extract information from the prediction procedure~\cite{BarredoArrieta2020}. In contrast, model-specific explanations are methods designed for certain models since they examine the internal model structures and parameters~\cite{du2019techniques}.
Fig.~\ref{fig:RelatedWork} depicts an overview of approaches for post-hoc explainability in PBPM.
Verenich et al.~\cite{verenich2019whitebox} propose a two-step decomposition-based approach. Their goal is to predict the remaining time. First, they predict on an activity-level the remaining time. Next, these predictions are aggregated on a process-instance-level using flow analysis techniques. 
Sindhgatta et al.~\cite{sindhgatta.2019} provide both global and local explainability for XGBoost, this is for outcome and remaining time predictions. 
Global explanations are on a prediction-model-level. Therefore, the authors implemented permutation feature importance. 
On the contrary, Local explanations are on a trace-level, i.e. they describe the predictions regarding a trace. For this, the authors apply LIME~\cite{ribeiro.2016}.
This method perturbs the input, observes how predictions change and based on that, tries to provide explainability.  
Mehdiyev and Fettke~\cite{Mehdiyev2020} present an approach to make DNN-based process outcome predictions explainable. Thereby, they generate partial dependence plots (PDP) to provide causal explanations.
Rehse et al.~\cite{rehse.2019} create global and local explanations for outcome predictions. Based on a DL architecture with LSTM layers, they apply a connection weight approach to calculate the importance of features and therefore provide global explainability. For local explanations, the authors determine the contribution to the prediction outcome via learned rules and individual features.

\vspace{-0.5cm}

In comparison to those approaches, LRP is not part of the training phase and presumes a learned model. LRP peaks into the model to calculate relevance backwards from the prediction to the input. Thus, through the use of LRP, we contribute by providing the first model-specific post-hoc explanations of LSTM-based next activity predictions.
\section{XNAP: Explainable Next Activity Prediction}
\label{sec:xnap}
\magic\ is composed of an offline and an online component.
In the offline component, a predictive model is learned from a historical event log by applying a Bi-LSTM DNN.
In the online component, the learned model is used for producing next activity predictions in running traces. Given the next activity predictions and the learned predictive model, LRP determines relevance values for each activity of running traces.

\subsection{Offline Component: Learning a Bi-LSTM model}
\label{sec:offline}
The offline component receives as input an event log, pre-processes it, and outputs a Bi-LSTM model learned based on the pre-processed event log.

\textbf{Pre-processing:}
The offline component's pre-processing step transforms an event log $L$ into a data tensor $\mathsf{X}$ and a label matrix $\mathbf{Y}$ (i.e. next activities). The procedure comprises four steps.
First, we transform an event log $L$ into a matrix $\mathbf{S} \in \mathbb{R}^{E \times U}$. $E$ is the event log's size $\vert L \vert$, whereas $U$ is the number of an event tuple's elements. Note that we add an activity to the end of each sequence to predict their end. 
Second, we onehot-encode the string values of the activity attribute in $\mathbf{S}$ because a Bi-LSTM requires a numerical input for calculating forward and backward propagations. After this step, we get the matrix $\mathbf{S} \in \mathbb{R}^{E \times H}$, where $H$ is the number of different activity values in the event log $L$.
Third, we create prefixes and next activity labels.
Thereby, a tuple of prefixes $R$ is created from $\mathbf{S} \in \mathbb{R}^{E \times H}$ by applying the function $f_{p}$, whereas a tuple of labels $K$ is created from $\mathbf{S} \in \mathbb{R}^{E \times H}$ through the function $f_{l}$.
Lastly, we construct a third-order data tensor $\mathsf{X} \in \mathbb{R}^{\vert R \vert \times M \times H}$ based on the prefix tuple $R$ as well as a label matrix $\mathbf{Y} \in \mathbb{R}^{\vert K \vert \times H}$ based on the label tuple $K$, where $M$ is the longest trace in the event log $L$, i.e. $\vert max_{\sigma}(L)\vert$. The remaining space for a sequence $ \sigma_{c}\in \mathsf{X}$ is padded with zeros, if $\vert \sigma_{c} \vert < \vert max_{\sigma}(L) \vert$.

\textbf{Model learning:}
\magic{} learns a Bi-LSTM model $\mathcal{M}$ that maps the prefixes onto the next activity labels based on the data tensor $\mathsf{X}$ and label matrix $\mathbf{Y}$ from the previous step. 
We use the Bi-LSTM architecture, an extension of ``vanilla" LSTMs since Bi-LSTMs are forward and backward LSTMs that can exploit control-flow information from two directions of sequences.
\magic's Bi-LSTM architecture comprises an input layer, a hidden layer, and an output layer. The \emph{input layer} receives the data tensor and transfers it to the hidden layer. 
The \emph{hidden layer} is a Bi-LSTM layer with a dimensionality of 100, i.e. the Bi-LSTM's cell internal elements have a size of 100. We assign the activation function $tanh$ to the Bi-LSTM's cell output. 
To prevent overfitting, we perform a random dropout of $20\%$ of input units along with their connections. %
The model connects the Bi-LSTM's cell output to the neurons of a dense \emph{output layer}.
Its number of neurons corresponds to the number of the next activity classes.
For learning weights and biases of the Bi-LSTM architecture, we apply the \emph{Nadam} optimisation algorithm %
with a \emph{categorical cross-entropy loss} and default values for parameters.
Note that the loss is calculated based on the Bi-LSTM's prediction and the next activity ground truth label stored in the label matrix $\mathbf{Y}$. 
Additionally, we set the batch size to $128$. Following Keskar et al.~\cite{keskar.2016}, gradients are updated after each 128\textsuperscript{th} trace of the data tensor $\mathsf{X}$. Larger batch sizes tend to sharp minima and impair generalisation. The number of epochs (learning iterations) is set to $100$, to ensure convergence of the loss function.

\subsection{Online Component: Producing predictions with explanations}
\label{sec:online}
The online component receives as input a running trace, performs a pre-processing, 
creates a next activity prediction and concludes with the creation of a relevance value for each activity of the running trace regarding the prediction.  
The prediction is obtained by using the learned Bi-LSTM model from the offline component.
Given the prediction, LRP determines the activity relevances by backwards passing the learned Bi-LSTM model.

\textbf{Pre-processing:}
The online component's pre-processing step transforms a running trace $\sigma_{r}$ into a data tensor and a label matrix, as already described in the offline component's pre-processing step. Note that we terminate the online phase if $\vert \sigma_{r} \vert$ is $\leq 1$ since, for such traces, there is insufficient data to base prediction and relevance creation upon.
Further, we assume that we have already observed all possible activities as well as the longest trace in the offline component. Thus, matrix $\mathbf{S}$ and tensor $\mathsf{X}$ lay in $\mathbb{R}^{1 \times H}$ and $\mathbb{R}^{1 \times M \times H}$. In the offline component, next activity labels are not known and based on the data tensor $\mathsf{X}_{r}$ for a running trace $\sigma_{r}$ a next activity is predicted.

\textbf{Prediction creation:} Given the data tensor $\mathsf{X}_{r}$ from the previous step, the trained Bi-LSTM model $\mathcal{M}$ from the offline component returns a probability distribution $\mathbf{p}^{1 \times H}$, containing the probability values of all activities. We retrieve the prediction $p$ from $\mathbf{p}$ through $argmax(\mathbf{p}[j])$, with $1 \leq j \leq H$.

\textbf{Relevance creation:} Lastly, we provide explainability of the prediction $p$ by applying LRP.  
For a next activity prediction $p$, LRP determines a relevance value for each activity in the course of a running trace $\sigma_{r}$ towards it by decomposing the prediction, from the output layer to the input layer, backwards through the model. Note the prediction $p$ was created in the previous step based on all activities of the running trace $\sigma_{r}$. 
In doing that, we apply the LRP approach proposed by Arras et al.~\cite{arras.2017} that is designed for LSTMs.  
As mentioned in Sec. \ref{sec:background}, a layer of a DNN can be described by one or more neural connections. Depending on the layer's type, LRP defines rules for attributing the relevance to lower-layer neurons given the relevance values of the upper-layer neurons.
After backwards passing the model by considering conversation rules of different layers, LRP returns a relevance value for each onehot-encoded input activity of the data tensor $\mathsf{X}$. Finally, to visualise the relevance values, e.g. by a heatmap, positive relevance values are rescaled to the range $[0.5, 1.0]$ and negative ones to the range $[0.0, 0.5]$.

\section{Results}
\label{sec:eval}
\subsection{Event logs}
We demonstrate the benefit of \magic{} with two real-life event logs that are detailed in Table~\ref{tab:eventlogs}.
\vspace{-0.7cm}

\begin{table}[htb]
\caption{Overview of used real-life event logs.}
\label{tab:eventlogs}
\resizebox{\textwidth}{!}{%
\begin{tabular}{|l|l|l|l|l|l|l|}
\hline
\textbf{Event log} & 
\textbf{\#instances} & \textbf{\begin{tabular}[c]{@{}l@{}}\# instance \\ variants\end{tabular}} & 
\textbf{\# events} & 
\textbf{\# activities} & \textbf{\begin{tabular}[c]{@{}l@{}}\# events \\ per instance$^*$\end{tabular}} & \textbf{\begin{tabular}[c]{@{}l@{}}\# activities \\ per instance$^*$\end{tabular}} \\ \hline
helpdesk           & 4,580                & 226                                                                      & 21,348             & 14                     & {[}2;15;5;4{]}                                                             & {[}2;9;4;4{]}                                                               \\ \hline
bpi2019           & 24,938               & 3,299                                                                    & 104,172            & 31                     & {[}1;167;4;4{]}                                                            & {[}1;11;4;4{]}                                                              \\ \hline
\multicolumn{7}{l}{} $^*$[\text{min; max; mean; median}]
\end{tabular}%
}
\end{table}

\vspace{-0.7cm}
First, we use the \textit{helpdesk} event log\footnote{https://data.mendeley.com/datasets/39bp3vv62t/1.}. It contains data of a ticketing management process form a software company.
Second, we make use of the \textit{bpi2019} event log\footnote{https://data.4tu.nl/repository/uuid:a7ce5c55-03a7-4583-b855-98b86e1a2b07.}. It was provided by a coatings and paint company and depicts an order handling process. Here, we only consider sequences of max. $250$ events and extract a 10\%-sample of the remaining sequences to lower computation effort.

\subsection{Experimental Setup}

LRP is a model-specific method that requires a trained model for calculating activity relevances to explain predictions. Therefore, we report the predictive quality of the trained models, and then demonstrate the activity relevances. 

\textbf{Predictive quality:} To improve model generalisation, we randomly shuffle the traces of each event log. For that, we perform a process-instance-based sampling to consider process-instance-affiliation of event log entries. This is important since LSTMs map sequences depending on the temporal order of their elements.
Afterwards, for each event log, we perform a ten-fold cross-validation. Thereby, in every iteration, an event log's traces are split alternately into a 90\%-training and 10\%-test set. Additionally, we use 10\% of the training set as a validation set.
While we train the models with the remaining training set, we use the validation set to avoid overfitting by applying early stopping after ten epochs. %
Consequently, the model with the lowest validation loss is selected
for testing.
To measure predictive quality, we calculate the average weighted \textit{Accuracy} (overall correctness of a model) and average weighted \textit{F1-Score} (harmonic mean of \emph{Precision} and \textit{Recall}). 

\textbf{Explainability:} To demonstrate the explainability of \magic's LRP, we pick the Bi-LSTM model with the highest \textit{F1-Score} value and randomly select two traces from all traces of each event log. One of these traces has a size of five; the other one has a size of eight. We use traces of different sizes to investigate our approach's robustness. 

\textbf{Technical details:}
We conducted all experiments on a workstation with 12 CPU cores, 128 GB RAM and a single GPU NVIDIA Quadro RXT6000. We implemented the experiments in \textit{Python} 3.7 with the DL library \textit{Keras}\footnote{https://keras.io.} 2.2.4 and the \textit{TensorFlow}\footnote{https://www.tensorflow.org.} 1.14.1 backend.
The source code can be found on Github\footnote{https://github.com/fau-is/xnap.}.
\subsection{Predictive Quality}
The Bi-LSTM model of \magic{} predicts the next most likely activities for the \textit{helpdesk} event log with an average (Avg) \textit{Accuracy} and \textit{F1-Score} of 84\% and 79.8\% (cf. Table~\ref{tab:resqual}). 
For the \textit{bpi2019} event log, the model achieves an Avg \textit{Accuracy} and \textit{F1-Score} of 75.5\% and 72.7\%. For each event log, the standard deviation (SD) of the \textit{Accuracy} and \textit{F1-Score} values is between 1.0\% and 1.5\%.  
\vspace{-0.7cm}
\begin{table}[htb]
\caption{Predictive quality of \magic's Bi-LSTM model for the 10 folds.}
\label{tab:resqual}
\resizebox{\textwidth}{!}{%
\begin{tabular}{|l|l|l|l|l|l|l|l|l|l|l|l|l|l|}
\hline
\textbf{Event log}                 & \textbf{Metric}   & 1 & 2 & 3 & 4               & 5      & 6      & 7               & 8      & 9      & 10               & \textbf{Avg} & \textbf{Sd} \\ \hline
\multirow{2}{*}{helpdesk} & 
\textit{Accuracy} & 
0.846	& 0.851 &	0.824 & 0.824 &	0.852 &	0.823 &	0.837 &	0.850 & 0.853 &	0.839
& 0.840       & 0.012      \\ \cline{2-14} 
& \textit{F1-Score} & 0.807 & 0.811	& 0.779 &	0.780 & 0.813 &	0.777 &	0.794 &	0.810	& \textbf{0.814} & 0.798 &	0.798	& 0.015\\ 
\cline{1-14}
\multirow{2}{*}{bpi2019}  & \textit{Accuracy} & 0.758     & 0.759     & 0.748     & 0.762          & 0.754  & 0.758 & 0.753          & 0.734 & 0.748 & 0.772          & 0.755       & 0.010      \\ \cline{2-14} 
& \textit{F1-Score} & 0.732     & 0.737      & 0.712     & 0.741          & 0.722 & 0.730 & 0.723          & 0.710 & 0.720   & \textbf{0.742} & 0.727       & 0.011 \\ 
\hline
\end{tabular}%
}
\end{table}
\vspace{-0.7cm}
\subsection{Explainability}
\begin{figure}[htpb]
\centering
\includegraphics[width=\textwidth]{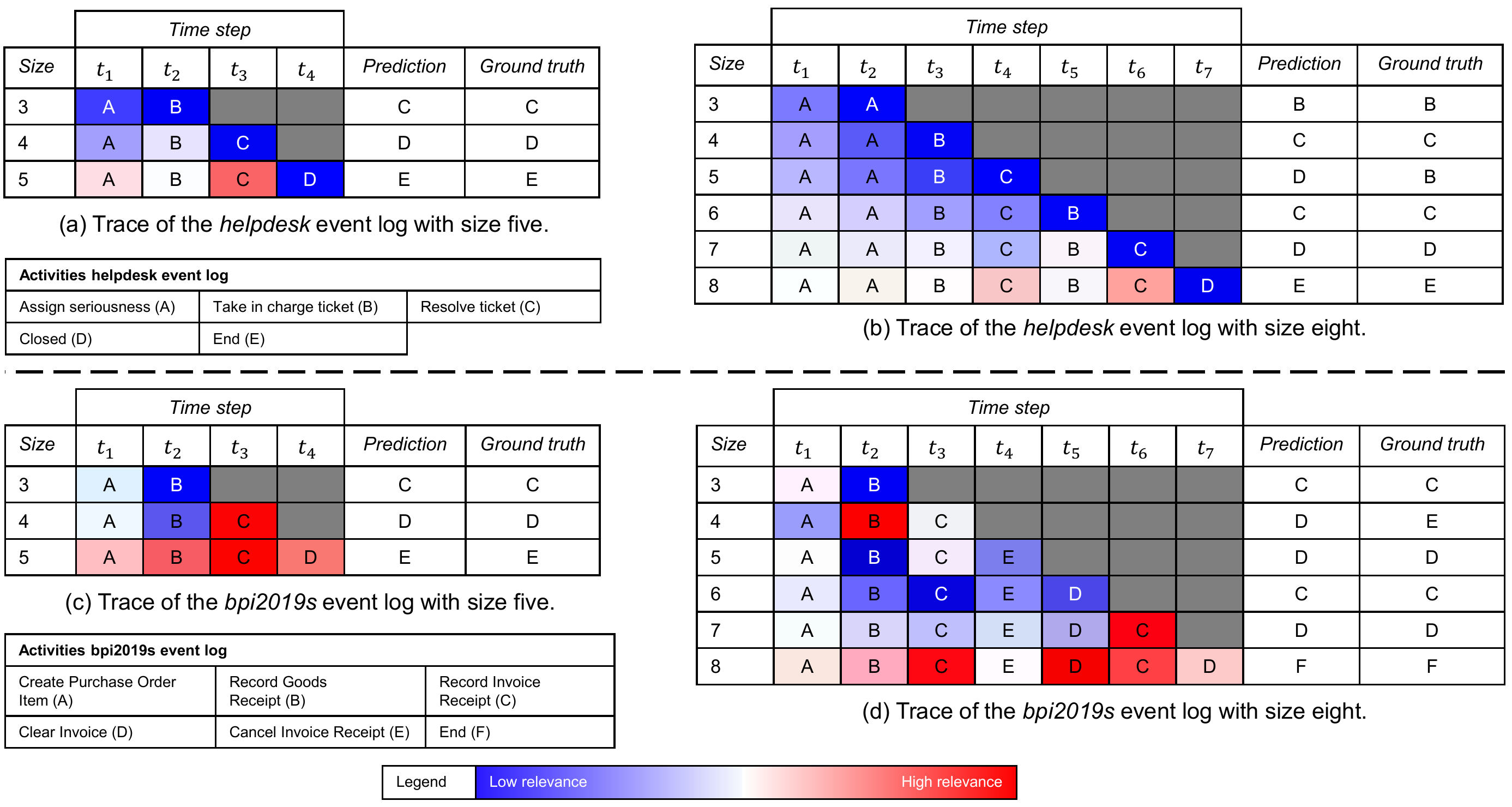}
\caption{Activity relevances of \magic's LRP.} 
\label{fig:expl}
\end{figure}

We show the activity relevance values of \magic's LRP on the example of two traces per event log (cf. Fig.~\ref{fig:expl}). The time steps (columns) represent the activities that are used as input. For each trace, we predict the next activity for different prefix lengths (rows). We start with a minimum of three and make one next activity prediction until the maximum length of the trace is reached (five and eight in our examples). The data-given ground truth is listed in the last column. We use a heatmap to indicate the relevance of the input activities to the prediction of the same row.
\vspace{-0.5cm}
\vspace{-0.5cm}
For example, in the traces (a) and (b), the activity ``Resolve ticket" (C) has a high relevance on predicting the next activity ``End (E)". With that, a process analyst knows that the trace will end since the ticket was resolved.
Another example is in the traces (c) and (d), where the activity ``Record Invoice Receipt (C)" has a high relevance on predicting the next activity ``Clear Invoice (D)". Thus, a process analyst knows that the invoice can be cleared in the next step because the invoice receipt was recorded.

\section{Conclusion}
\label{sec:conclusion}
Given the fact that DNNs achieve a promising predictive quality at the expense of explainability and based on our identified research gap, we argue that there is a crucial need for making LSTM-based next activity predictions explainable.  
We introduced \magic{}, an explainable PBPM technique, that integrates an LRP method from the field of XAI to make a BI-LSTM's next activity prediction explainable by providing a relevance value for each activity in the course of a running trace.  
We demonstrated the benefits of \magic{} with two event logs. By analysing the results, we made three main observations.
First, LRP is a model-specific XAI method; thus, the quality of the relevance scores depend strongly on the model's predictive quality. 
Second, \magic{} performs better for traces with a smaller size and a higher number of different activities.
Third, \magic{} computes the relevance values of activities in very few seconds. In contrast, model-agnostic approaches, e.g. PDP~\cite{Mehdiyev2020}, need more computation time.

In future work, we plan to validate our observations with further event logs. Additionally, we will conduct an empirical study to evaluate the usefulness of \magic{}. We also plan on hosting a workshop with process analysts to better understand how a prediction's explainability contributes to the adoption of a PBPM system.
Moreover, we plan to adapt the propagation rules of \magic's LRP also to determine relevance values of context attributes.
Another avenue for future research is to compare the explanation capability of a model-specific method like 
LRP to a model-agnostic method like LIME for, e.g. the DNN-based next activity prediction.
Finally, \magic's explanations, which are rather simple, might not capture an LSTM model's complexity. Therefore, future research should investigate new types of explanations that better represent this high complexity.

\section*{Acknowledgments}
This project is funded by the German Federal Ministry of Education and Research (BMBF) within the framework programme \textit{Software Campus} under the number 01IS17045. The fourth author received a grand from \"Osterreichische Akademie der Wissenschaften.

 \bibliographystyle{splncs04}
 \bibliography{bpm}
\end{document}